# Speaker Independent Continuous Speech to Text Converter for Mobile Application

*R.Sandanalakshmi, P.Abinaya viji, M.Kiruthiga, M.Manjari, A.Sharina
Department of Electronics and communication |Engineering
Pondicherry Engineering College
Puducherry, India
*sandanalakshmi@pec.edu

**Abstract:**

An efficient speech to text converter for mobile application is presented in this work. The prime motive is to formulate a system which would give optimum performance in terms of complexity, accuracy, delay and memory requirements for mobile environment. The speech to text converter consists of two stages namely front-end analysis and pattern recognition. The front end analysis involves preprocessing and feature extraction. The traditional voice activity detection algorithms which track only energy cannot successfully identify potential speech from input because the unwanted part of the speech also has some energy and appears to be speech. In the proposed system, VAD that calculates energy of high frequency part separately as zero crossing rate to differentiate noise from speech is used. Mel Frequency Cepstral Coefficient (MFCC) is used as feature extraction method and Generalized Regression Neural Network is used as recognizer. MFCC provides low word error rate and better feature extraction. Neural Network improves the accuracy. Thus a small database containing all possible syllable pronunciation of the user is sufficient to give recognition accuracy closer to 100%. Thus the proposed technique entertains realization of real time speaker independent applications like mobile phones, PDAs etc.

## 1. INTRODUCTION

**Speech recognition** (SR) is the translation of spoken words into text. It is also known as "automatic speech recognition", "ASR", "computer speech recognition", "speech to text", or "STT". Speech is a natural mode of communication for people and it conveys some information. This information can be used for different purposes like authentication, text conversion or machine control depending upon application. People feel so comfortable to interact with computers via speech, rather than resorting to primitive interfaces such as keyboards and pointing devices. The need for speaker independent continuous speech to conversion system lies at the core of many rapidly growing application areas. A *speaker independent* system is intended for use by any speaker. C*ontinuous speech* means naturally spoken sentences, separated by minimum silence which is used for detecting boundaries. Continuous speech recognition is difficult when compared to Isolated words speech recognition. A speech interface would support many valuable applications like telephone directory assistance, spoken database querying for novice users, "handsbusy" applications in medicine or fieldwork, office dictation devices and for controlling electronic devices. Especially, it is useful for embedded systems like smart phones and PDAs having insufficient space for typing or touching and helpful for controlling navigation during car driving. Also, it can be used to build advanced security systems and ATM machines.

At present, there have been a number of successful commercial voice interfaces. The most prominent example is Siri, the voice-activated personal assistant built in the latest iphone. Speech recognition products are also available in Android, the Windows Phone platform, and most other mobile systems with considerable limitations. The recognition accuracy and performance of a system would degrade dramatically with small modifications of speech signal or speaking environment. As a result more computation and memory capacity are needed for Speech recognition.

Speech acquisition is the first and foremost stage of speech recognition. It is a process of acquiring raw speech signals from user. The acquired speech needs to be preprocessed before extracting information. The extracted information is used for recognition. The most important step in preprocessing is speech detection in the acquired signal. The effectiveness of speech recognizer is crucially dependent on its performance. This is an algorithm to detect silence parts of a speech signal and remove it as it does not provide any information. Speech pause detection algorithm [1] is used which detects speech pause minima by adaptively tracking minima in a noisy signal power envelope. An ideal voice activity detector needs to be independent from application area and noise condition. But at the same time, the VAD algorithm [2] should be of low complex to facilitate real time application. Therefore simplicity and robustness against noise are two essential characteristics of a practicable voice activity detector which is incorporated in the proposed system. The traditional voice activity detection algorithms which track only energy cannot successfully identify potential speech from input. VAD should be most sensitive as the unwanted part of the speech also has some energy and appears to be speech. This demands to use a VAD that also calculates energy of high frequency part separately as ZCR to differentiate noise from speech [3].

The preprocessed signal needs to be represented in a compact manner for further process. Speech carries much linguistic information which is required for recognition task. This feature extraction technique is based on Mel scale. MFCC [4] overcomes this limitation. Vocal tract including tongue, teeth etc. determines the sound generated by the human. The shape of the vocal tract determines what sound comes out. If we are able to determine this shape, accurate representation of the speech sample can be produced. The shape of the vocal tract manifests itself in the envelope of the short time power spectrum, and the job of MFCCs is to accurately represent this envelope. MFCC is proved to be more efficient by comparing MFCC with other feature extraction techniques.[5]. In recent years Neural Network is beginning to gain importance. This paved a path for efficient deployment of recognizer. The usability of neural network for speech recognizer proved to be a suitable technology [6] and it outperforms traditional recognizers like Hidden Markov Model (HMM), Dynamic Time Warping (DTW), Vector Quantization (VQ) etc. The promising results were obtained using Radial Basis Neural Network as recognizer [7]. It also is found that RBF trains and tests faster than Multilayer Perceptron (MLP) Neural Networks. Moreover RBF gives more accurate result than MLP [8]. In the proposed system GRNN [9] is used as recognizer. A Generalized Regression Neural Network is often used for function approximation. It is similar to Radial Basis Neural Network but has a special linear layer in addition to RBF network. It is found that the performance of GRNN is superior to the other classifiers namely Linear and MLP Neural Networks [10].

## 2. PREPROCESSING

### 2.1 SPEECH ACQUISITION

An acoustic speech signal exists as pressure variations in the air. A microphone converts these pressure variations into an electric current. To process the speech signal digitally, it is necessary to make the analog waveform discrete in both time (sample) and amplitude (quantize). This A-to-D conversion is generally accomplished by digital signal processing hardware on the computer's sound card. The spoken voice frequency lies between 300 to 3400 Hertz. So a sampling frequency of 8 KHz is chosen to satisfy Nyquist criterion. The speech signal is recorded with a sampling rate of 8 KHz.. And the data is saved with distinct name with '.wav' as extension.

### 2.2 VOICE ACTIVITY DETECTION

Voice activity detection is a technique to detect the un-silenced part of the incoming speech signal. In general, the acquired speech signal of 1 second duration (8000 samples) starts and ends with silence which accounts for nearly 6000 samples. VAD can be performed by two algorithms. The first algorithm uses signal features based on energy level and the second algorithm uses signal features based on the rate of zero crossings. The combination of both gives good result that is used in the proposed system. A basic VAD works on the principle of extracting measured features from the incoming audio signal, which is divided into frame size of 150 ms duration and frame increment size of 40 ms duration. The extracted signal features based on energy level and zero crossing rate from the audio signal are then compared to a threshold and then VAD decision is computed.

STEP 1: If the feature of the input frame exceed the estimated threshold value, a VAD decision (VAD = 1) is computed which declares that speech is present.

STEP 2: Otherwise, a VAD decision (VAD = 0) is computed which declares the absence of speech in the input frame.

### 2.3 PRE-EMPHASIS

In order to flatten speech spectrum, a pre-emphasis filter is used before spectral analysis. Its aim is to compensate the high-frequency part of the speech signal that was suppressed during the human sound production mechanism. Thus high frequency signals have less amplitude. This gives rise to a negative spectral slope and is compensated by appropriate pre-emphasis filter. The z-transform of the filter is given by [2]:

$$H(z) = 1 - az^{-1}, \ 0.9 \leq a \leq 1.0 \qquad (1)$$

Where 'a' is the filter coefficient and is chosen as 0.9375 The output signal after pre-emphasis is given by [2]:

$$y(n) = s(n) - a*s(n-1) \qquad (2)$$

Where y(n) - output signal
s(n) - input speech signal

## 2.4 FRAMING

Speech is a highly non-stationary signal. Hence speech analysis must be carried out on short segments across which the speech signal is assumed to be stationary. For this the speech signal is divided into frames of small duration typically 20 to 40ms with overlap of 10 to 15ms for short-time spectral analysis.

## 2.5 WINDOWING

Windowing minimizes the discontinuities by tapering the signals to quite small values (nearly zeros) at the edges of a frame. Hamming window is also called the raised cosine window. The Hamming window defined as [3]

$$w(n) = 0.54 - 0.46 \cos(2\pi n), \quad 0 \leq n \leq N \quad (3)$$
$$= 0, \quad \text{otherwise}$$

The output speech signal can be described as

$$x(n) = w(n).s(n), \quad 0 \leq n \leq N \quad (4)$$

Each frame s(n) is multiplied by hamming window w(n) of 25ms duration.

## 3. FEATURE EXTRACTION

Feature extraction is the process of distinguishing features from the input signal. It is about reducing the dimensionality of the input-vector while still maintaining the uniqueness of the signal. Good features possess the attributes of being simple to extract, invariant against time translations and overall amplitude variation, and capable of discriminating different phonetic categories. Features are extracted using MFCC technique [4].

## 3.1 MEL-FREQUENCY CEPSTRAL COEFFICIENT ALGORITHM

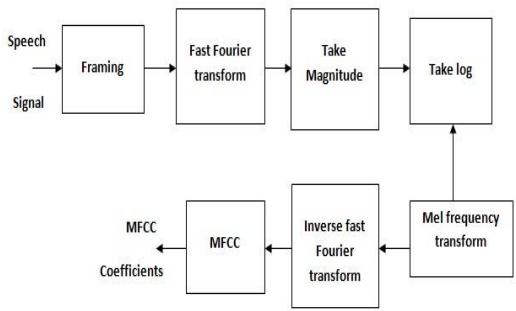

**Fig. 1 MFCC block diagram**

MFCCs are commonly derived as follows:

- On short time scales the audio signal doesn't change much. The frame time chosen is 20-40ms frames.

- The power spectrum of each frame is passed through the FFT block. FFT is used for identifying the frequencies present in the frame.

- The logarithmic compression is performed to match features more closely to what humans actually hear

- The periodogram spectral estimate still contains a lot of information not required for Automatic Speech Recognition (ASR). For this reason we take clumps of periodogram bins and sum them up to get an idea of how much energy exists in various frequency regions. This is performed by our Mel filter bank. The Mel scale tells us exactly how to space our filter banks and how wide to make them.

- The inverse fast Fourier transform of the Mel log power gives the MFCC coefficients.

## 4. SPEECH RECOGNIZER USING NEURAL NETWORK

The role of a recognizer is to assign the feature vector provided by the feature extractor to a category. It aims at measuring the similarity between an input speech and a reference pattern or model which is obtained during training. A neural network is a computational program that is designed to model the way in which the brain performs a particular task or function of interest; the network is usually implemented using electronic components or simulated in software on a digital computer.

### 4.1 PROPOSED NN RECOGNIZER

The goal of speech regression analysis is to model a recognizer by providing a finite set of observations of speech and its associated target values such that it gives most probable output value for an unknown speech input. The regression equation can be expressed as [9]

$$y_j = \frac{\sum_{i=1}^{n} w_{ij} h_i}{\sum_{i=1}^{n} h_i} \quad (5)$$

$$h_i = \exp\left(-\frac{D_i^2}{2\sigma^2}\right) \quad (6)$$

where $w_{ij}$ is the target output corresponding to input training vector $x_i$ and $j^{th}$ output.

The above equations (5) and (6) can be implemented as the following neural network structure.

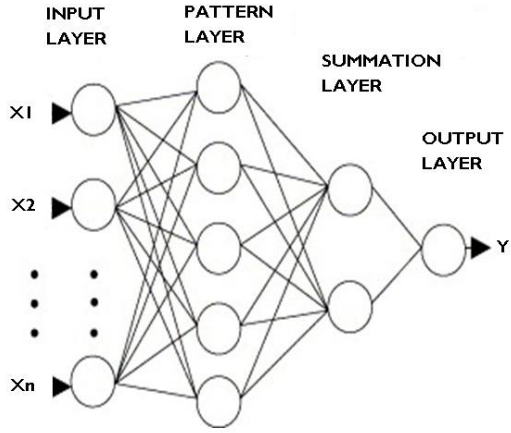

Fig. 2 Regression neural network architecture

- The input layer has neurons which represent the input patterns. It passes the input vector to each of the neuron in the pattern layer.
- The pattern layer has one neuron for each pattern. During training all variations of input vectors and its associated desired output or target ($w_{ij}$) will be provided to the network. The neuron in the pattern layer compute $C_i$ for the input provided to it.

During testing the neuron will compute $h_i$ using $C_i$ and $X_i$ as shown below[9]:

$$D_i^2 = \| C_i - X_i \| \quad (7)$$
$$h_i = \exp\left(-\frac{D_i^2}{2\sigma^2}\right) \quad (8)$$

where $\sigma$ is the spread (width) of the kernel function.

Spread denotes the distance an input vector must be from a neuron's $C_i$ in order to be correctly classified. The value of spread chosen for the implemented network is 0.5. The summation layer has two units N and D where N computes $\sum_{i=1}^{n} w_{ij} h_i$ and D calculates $\sum_{i=1}^{n} h_i$.

The output unit divides N by D which gives the prediction result. The above implementation is illustrated in the figure shown below:

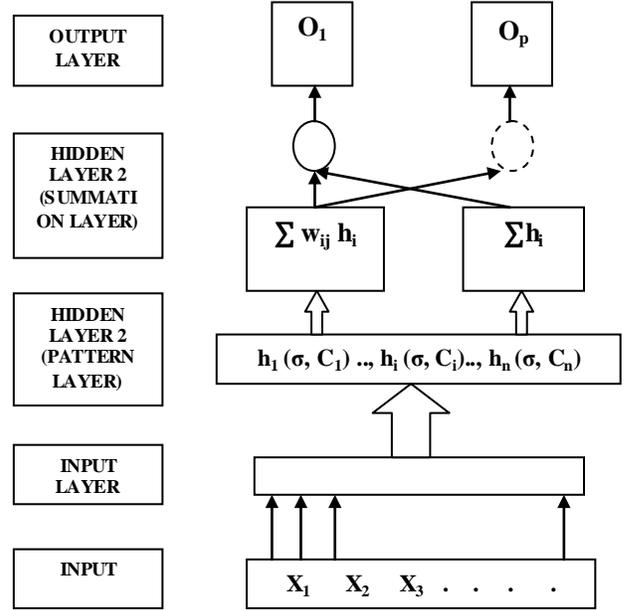

Fig. 3 GRNN recognizer

The above neural network is known as General Regression Neural Network (GRNN). It is one of the radial basis networks. The hidden layer neurons of this network are referred as radial basis neurons as they employ radial basis kernel function. The proposed network does superior function approximation and thus is well suited for speech recognition.

## 5. DATABASE CREATION

In order to facilitate the training and testing of the recognizer, speech database is required. A variety of speech samples were obtained from different speakers to form the speech database. Three kinds of databases were created for

- Digit Recognition
- Word recognition
- Transcription of title

### 5.1 Digit Recognition:

The digit database consists of ten utterances 'zero' to 'nine' collected from both male and female speakers. The database is divided into training and testing. Training set is used to train the neural network. Testing set is used to test the performance of neural network. For speaker dependent recognition, excellent recognition accuracy closer to 100% is achieved. This recognizer can be very well adapted for voice dialing.

## 6. RESULTS AND DISCUSSION

### 6.1 Syllable Level Word Recognition:

To perform syllable based implementation of word recognition, a separate database was created with speech samples from 2 female speakers. 36 syllables, namely, 'be', 'com', 'in', 'out', 'ply', 'pose', 'press', 'side', 'sup', 'in', 'vi', 'ta', 'tion', 'ti', 'ma', 'no', 're', 'po', 'si', 'lo', 'de', 'di', 'ca, 'car', 'na', 'mo', 'go', 'to', 'lo', 'la', 'ra', 'ing', 'vo', 're', 'va', 'do' were selected. 4 samples were taken for each word. The number of words in a language is numerous. Training a network for each word requires huge memory which is not realistic. To reduce memory requirement, language structure can be exploited. The number of syllables in a language is comparatively much lesser than that of words. A word is formed by the combination of one or more syllables and many words could be built up from a single syllable. This idea has lead to training of another GRNN recognizer for syllables instead of complete words. The recognizer was designed with 36 syllables from which 80 plus words could be formed. The testing of the recognizer gave heartening results.

### 6.2 Transcription:

Transcription is a process of converting continuous speech into text. For real-time scenario, speech signal for recognition would never be an isolated word but rather a continuous stream of words. Syllable level recognition with minimum pauses can also extended for transcription purposes. To perform transcription of the title "SPEAKER INDEPENDENT CONTINUOUS SPEECH TO TEXT CONVERSION SYSTEM" a separate database was created, consisting of each word. The samples were collected from both male and female speakers.

**Table.2 Recognition Accuracy**

| Speaker Details | Accuracy* |
|---|---|
| Trained speakers | 96% |
| Untrained speakers | 85% |

\* Accuracy = Number of words correctly recognized / Total number of words

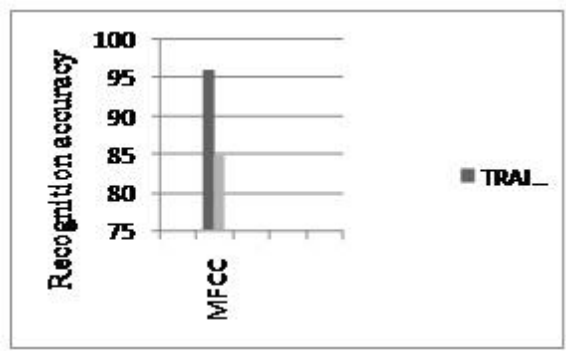

Fig. 4 Comparison chart of Recognizer performance using MFCC

| TYPES | Correlation between | | | | Elapsed time (seconds) |
|---|---|---|---|---|---|
| | Same samples from different speakers | Similar sounding samples | Distinct samples | Noised sample | |
| RASTAPLP | 0.8360 | 0.8415 | 0.6310 | 0.6998 | 0.5587 |
| MFCC | 0.9599 | 0.7268 | 0.6548 | 0.8285 | 0.8246 |
| BFCC | 0.9762 | 0.9281 | 0.7962 | 0.9737 | 0.7683 |
| RPLP | 0.9922 | 0.9154 | 0.7341 | 0.9389 | 0.9079 |
| MFPLP | 0.9793 | 0.8283 | 0.6679 | 0.7195 | 0.8987 |
| PLP | 0.9808 | 0.9390 | 0.8075 | 0.9806 | 0.8837 |

Table 1 Comparison table of different Feature Extraction Techniques

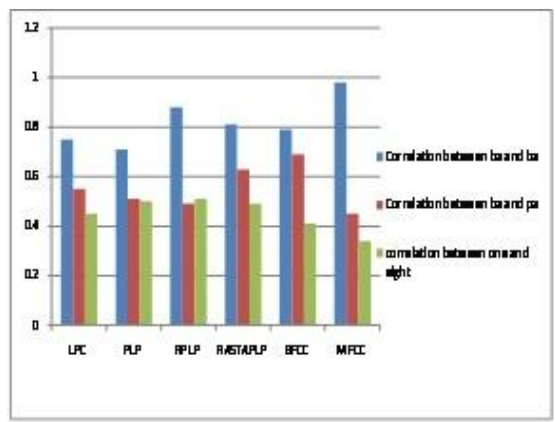

Fig.5 Comparison chart for different Feature Extraction Techniques

➢ Comparison chart for elapsed time

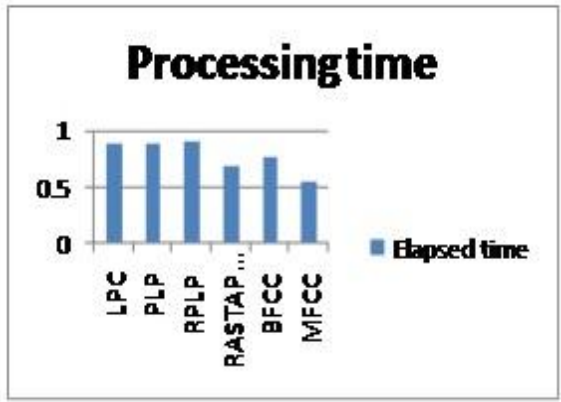

**Fig 6 Comparison chart for different Feature Extraction Techniques**

The table consisting the correlation values between two same words ( ba and ba ), between two similar sounding words ( ba and pa), between two distinct word(one and eight), elapsed time and correlation comparison with noise is shown. The correlation analysis has been performed on RASTAPLP, MFCC, BFCC, RPLP, MFPLP, PLP features. From the performance chart given above it is obvious that the MFCC technique gives well distinct features than other techniques.

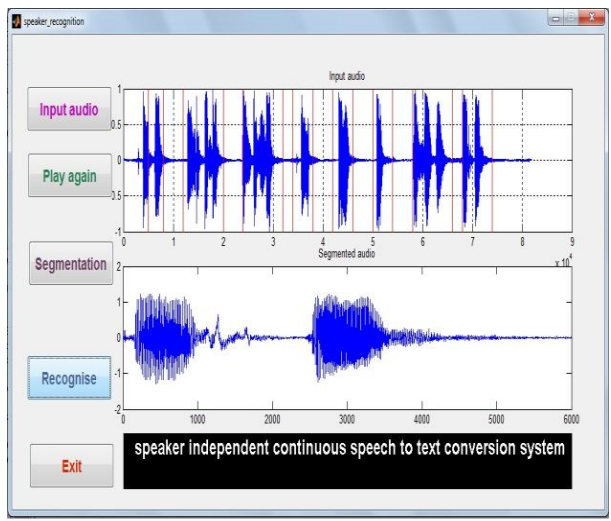

**Fig. 7 Screenshot of Graphical User Interface**

## 7. SUMMARY AND CONCLUSION

In this work, a continuous speech to text conversion system has been modeled using VAD, MFCC and GRNN network.. The proposed model has several advantageous characteristics such as fast learning, flexible network size and robustness to speaker variability (ability to recognize the same words pronounced in various manners). GRNN promises to be a successful and powerful alternative to the conventional speech recognizers.

The designed recognition system with all the above salient features has been utilized for developing a digit recognition system, a syllable level word recognition system and a transcription system. Digit recognition can find applications in voice dialing systems. And it needs to be a speaker independent system. In syllable level word recognition system, the recognizer was trained for syllables. These syllables were separately recognized and concatenated to output the complete word as text. In Transcription system, the duration of an input speech signal is made longer and the corresponding text output is obtained. The recognizer was trained for individual words. This system acts as a basis for real time speech recognition products, where input will never be a single word. Good recognition accuracy has been achieved in both cases. These implementations illustrate the potential of optimal configurations of key ASR components.

Theoretically, the accuracy increases with the increase in training data. As a result, memory needed also increases. It is found that carefully forming the database helps a lot in reducing memory requirements and increases recognition accuracy. In training phase, the words are spoken clearly so that it avoids general variations and confusions. In testing phase, the speech signal with minimum pauses should be given as an input. This enables the recognizer to discriminate the words effectively. For future work ,the language modeling of HMM can also be utilized in neural network implementation to build an efficient hybrid HMM-NN recognizer including better acoustic modeling accuracy, better context sensitivity, more natural discrimination and a more economical use of parameters.

Dr.R.Sandanalakshmi, is currently working as Assisstant Professor Dept. of Electronics and communication Engineering, Pondicherry Engineering College. She has total 12 years of teaching experience. Her research field of interest includes QoS improvement for next generation wireless networks, Non-invasive studies on prognosis of dengue using signal processing methods, speech to text conversion for mobile applications .

P.Abinaya viji, M.Kiruthiga, M.Manjari, A.Sharina completed Under Graduate B.Tech in the department of Electronics and Communication Engineering, Pondicherry Engineering College in April 2013 and placed in Core companies of specialization.